\documentclass[11pt, twocolumn]{article}

\usepackage[letterpaper, left=2.5cm, right=2.5cm, bottom=2cm, top=2cm]{geometry}

\usepackage{graphicx}
\usepackage{amsfonts}
\usepackage{amsmath}
\usepackage{booktabs}
\usepackage{float}
\usepackage{hyperref}
\usepackage{multicol}

\usepackage[brazilian]{babel}
\usepackage[utf8]{inputenc}
\usepackage[T1]{fontenc}

\usepackage{adjustbox}

\AtBeginDocument{%
\providecommand\BibTeX{{%
\normalfont B\kern-0.5em{\scshape i\kern-0.25em b}\kern-0.8em\TeX}}}

\AtBeginDocument{}

\begin{document}
\pagenumbering{gobble}

%\title{\line(1,0){450}\\\textbf{\texttt{Electricity Theft Detection with self-attention}}\\ \line(1,0){450}}

\title{\textbf{Normalizador Neural de Datas e Endereços}}

\author{
	\textbf{~Gustavo Plensack} ~ e-mail: g155662@g.unicamp.br\footnote{Os autores contribuiram igualmente a essa pesquisa}   
	\and \textbf{~Paulo Finardi} ~~~~~~~ e-mail: p144809@g.unicamp.br\footnotemark[\value{footnote}] 
}

\date{}

\maketitle
%\begin{multicols}{2}

\begin{abstract}
\textit{Documentos de qualquer natureza apresentam uma grande variedade de formatos de datas e endereços, em alguns casos datas podem estar escritas totalmente por extenso ou ainda possuir diferentes tipos de separadores. A desordem de padrão em endereços é ainda maior devido a maior possibilidade de permutação entre ruas, bairros, cidades e estados. No contexto de processamento de linguagem natural, problemas dessa natureza são tratados por ferramentas rígidas tais como ReGex ou DateParser, que são eficientes desde que a entrada esperada esteja pré-configurada. Quando esses algoritmos recebem um formato não esperado, erros e saídas indesejadas acontecem. Para contornar esse desafio, apresentamos uma solução com redes neurais profundas $estado~da~arte$, o \texttt{T5} que trata formatos não pré-configurados de datas e endereços com acurária acima de 90\% em alguns casos. Com esse modelo, nossa proposta trás generalização à tarefa de normalização de datas e endereços. Também tratamos esse problema com com um dado ruidoso que simula possíveis erros no texto.}
\end{abstract}

%\keywords{Attention Mechanisms, Fraud Detection, Deep Learning, Electricity-Theft Detection, Missing Data treatment, Smart-Grids}

\section{Introdução}
Lidar com formatos variados de datas e endereços é uma tarefa rotineira para qualquer ser humano. Somos capazes de entender diversos formatos de datas e endereços de forma trivial. No entanto tal atividade tratada por computador pode ser complexa. A complexidade vem do fato da grande variedade de formas que essas entidades podem possuir. O caminho mais tradicional para essa tarefa é o uso de técnicas clássicas de Processamento de Linguagem Natural (PLN), em especial o uso do \textit{Regular Expression} (RegEx) que é amplamente aplicado em normalizações de textos e inclusão/exclusão de caracteres. Outras ferramentas como o Dateparser \cite{Dateparser}  são empregadas no tratamento de normalização de datas e apresentam um bom desempenho. Esses métodos de normalização são baseados em estruturas rígidas e apresentam dois problemas: não existe generalização da solução, isto é, se existir um formato não previsto a priori o método falhará. Possuem uma tolerância pobre a dados ruidosos. Este artigo apresenta uma solução eficiente que trata esses dois problemas. Nossa proposta é utilizar um modelo de rede neural profunda estado da arte em PLN.

Nos últimos anos, em particular depois do desenvolvimento da arquiterura de rede neural do transformer \cite{Vaswani_2018}, as técnicas de PLN, em particular modelos de auto atenção  ajudaram no avanço de aprendizado profundo em outras áreas: problemas com dados estruturados em detecção de fraude elétrica \cite{fraude_elet} e problemas de visão computacional \cite{GoogleAtt}. O grande avanço das técnicas de PLN corrobora para o desenvolvimento prático de soluções em diversos nichos. Neste artigo, treinamos o modelo ~ \texttt{T5} \cite{T5} para lidar com tais situações. Em essência desenvolvemos dois modelos de normalização: um para datas e outro para endereços. Para o treinamento desses modelos, criamos dois datasets sintéticos, disponíveis neste repositório do \href{https://github.com/dl4nlp-rg/PF08-Gustavo-Paulo}{GitHub}.

O texto é apresentado da seguinte forma: a seção \ref{sec:1} apresenta trabalhos relacionados, a seção  \ref{sec:2} aborda a metodologia, a seção \ref{sec:dataset} detalha como os datasets de endereço e data foram construídos. Os resultados dos experimentos são mostrados na seção \ref{sec:3}. A conclusão e próximos desafios estão nas seções \ref{sec:4} e \ref{sec:5} respectivamente.

\section{Pesquisas relacionadas}
\label{sec:1}
Normalização de texto é uma das tarefas mais tradicionais em PLN. O artigo \cite{text_normalization} realiza normalização de texto com um modelo que utiliza redes neurais recorrentes (RNNs), entretanto o principal foco dessa pesquisa é transformar as palavras escritas normalizadas em uma forma audível, falada. Em \cite{hal} é apresentado uma normalização lexical de textos com o uso do BERT \cite{bert} que  enquadra a normalização lexical como uma tarefa de previsão de token. O trabalho descrito em \cite{sproat_rnn}  é o mais relacionado ao nosso projeto. Nele os autores treinaram um modelo de RNN em um grande dataset que contém formas de texto falada com o objetivo do modelo aprender a correta normalização do texto, entretanto os autores utilizaram filtros na saída da rede neural para obter melhores resultados. 

A natureza do problema de normalização de datas e endereços envolve uma grande coleção de padrões de reconhecimento, de modo que é inviável construir um método sem aprendizado que cubra todos as possibilidades. É justamente nesse ponto onde os algoritmos de aprendizado de máquina profundo são úteis devido à sua capacidade de generalizar para padrões não observados na etapa de treinamento.

\section{Metodologia}
\label{sec:2}
Nessa pesquisa estudamos um modelo de PLN que pode ser aplicado em etapas de pós-processamento na tarefa de leitura automática de documentos \cite{automatic_reading}.
Uma vez identificado o texto de uma data ou endereço, este será fornecido como entrada para o modelo estudado que deve deve retornar o formato canônico deste texto.
A avaliação desempenho será feita através do acurácia com correspondência exata entre a string gerada pelo modelo e o valor esperado.

\subsection{Normalização de Datas}
Desenvolvemos neste artigo três tipos de normalizações para datas:
\begin{itemize}
    \item Datas Completas: DD/MM/AAAA
    \item Datas Incompletas: DD/MM ou MM/AAAA
    \item Datas Relativas: há $100$ dias $\rightarrow$ $~~-100d$
\end{itemize}

\subsubsection{Datas Completas}
A saída esperada da normalização de datas completas tem a forma: DD/MM/AAAA. Com $45$ diferentes formatos e  $73000$ amostras, criamos um dataset com um intervalo de $200$ anos, entre $1921$ até $2120$. O conjunto de treino possui $34$ diferentes formatos e o conjunto de teste os $11$ formatos remanescentes.

\subsubsection{Datas Incompletas}
A saída esperada da normalização de datas incompletas possui o seguinte formato: DD/MM ou MM/AAAA. Para DD/MM, o dataset possui $2500$ amostras, para MM/AAAA, o dataset possui $7200$ amostras. Os $45$ diferentes formatos de amostras em ambos os casos também foram dividios em $34$ amostras para treino e $11$ para testes.

\subsubsection{Datas Relativas}
A saída esperada da normalização de datas relativas possuem as seguintes formas: para datas futuras em dias, meses ou anos, o número $N$ de dias, meses ou anos é seguido da letra $d-$dias, $m-$dias, ou $a-$anos. Para datas no passado é acrescentado o sinal de negativo. Com mais detalhes, os formatos são:
\begin{enumerate}
    \item $N$ dias $\rightarrow$ futuro $(+)$ ou passado $(-)$: $\pm Nd$
    \item $N$ meses $\rightarrow$ futuro $(+)$ ou passado $(-)$: $\pm Nm$
    \item $N$ anos $\rightarrow$ futuro $(+)$ ou passado $(-)$: $\pm Na$
\end{enumerate}
Para experimentos mais robustos, variamos o tamanho do dataset em 3 tamanhos. O dataset de datas relativas possui $18$ tipos diferentes de formatos, onde $5$ formatos são escolhidos aleatoriamente e utilizados somente no conjunto de testes, também existe a opção de escolha de lingua: Português ou Inglês e nível de ruído.

\subsection{Normalizador de Endereços}
O rótulo do experimentos de endereço possui a seguinte estrutura: \textit{logradouro, número, complemento, bairro, cidade e estado}. Os variados tipos do dataset incluem formatos de permutação dessa ordem, ruas e avenidas escritas com a palavra logradouro antes do nome e outros tipos. Os experimentos dessa solução são mostrados na tabela \ref{address}. 

\subsection{Ruído}
Uma das grandes vantagens propostas para a utilização de métodos neurais para normalização é a possibilidade lidar com ruídos, isto é, valores inesperados que podem ocorrer devido à erros humanos ou de tecnologias como OCRs. Tendo em mente estas possibilidades, durante o treinamento o modelo será apresentado aos seguintes formatos de ruídos:

\begin{itemize}
    \item troca de caracteres semelhantes\footnote{Alguns caracteres que podem ser semelhantes em algumas fontes são: $o\leftrightarrow 0~$, $c\leftrightarrow$ ç~, $l\leftrightarrow i~$,  $n\leftrightarrow m~$,  $u\leftrightarrow v~$,  $9\leftrightarrow g~$}.
    \item remoção de caracteres aleatórios.
    \item inclusão de caracteres aleatórios.
    \item quebra inesperada de palavras.
\end{itemize}

Para o caso dos endereços, também foram considerados ruídos abreviações comuns, como é o caso de:

\begin{itemize}
    \item $Rua \leftrightarrow R.$
    \item $Avenida \leftrightarrow Av.$
    \item $Santo \leftrightarrow Sto.$
\end{itemize}

\subsection{Normalização Unificada}
Levando em conta que o treinamento do T5 descrito em \cite{T5} foi feito considerando múltiplas tarefas, iremos avaliar também a capacidade deste modelo de normalizar endereços e datas com uma mesma instância do modelo, considerando a inclusão de prefixos ou não.

Usando esta metodologia, propomos avaliar a capacidade de genaralização do normalizador neural de datas e endereços, sua imunidade à ruído e também sua capacidade de normalizar datas e endereços de forma unificada.

\section{Datasets}
\label{sec:dataset}
Um ponto chave da nossa proposta é a criação dos datasets. Existem algumas dificuldades em encontrar bons conjuntos de dados para o problema de normalização de datas e endereços, pois é necessário que os dados possuam grande variedade de formatos e estejam em português. Sendo assim, a  abordagem adotada por nós neste trabalho foi criar datasets sintéticos para datas e endereços.

\subsection{Dataset para Datas}
A criação do dataset de datas, foi realizado a partir de formatos encontrados nas seguintes fontes:
\begin{itemize}
    \item DateParser \cite{Dateparser}; 
    \item LexNLP \cite{lexnlp};
    \item Wikipédia e portais de notícia;
\end{itemize}

O ponto mais significativo na criação deste dataset é a variabilidade. Este fator introduz dificuldade ao modelo de  prestar muita atenção aos dados do conjunto de treino, de modo a evitar a memorização dos formatos vistos em tempo de treinamento e melhorar a capacidade de generalizaçao do modelo.

Ao todo o dataset possui $45$ formatos para datas absolutas, $90$ para datas incompletas e $36$ para datas relativas para o português brasileiro. No caso da língua inglesa, foram considerados $45$ formatos para datas absolutas, $90$ para datas incompletas e $18$ para datas relativas. Estes formatos variam entre si alterando os separadores, forma de escrita dos números e preposições no caso das datas relativas.

A implementação do gerador do dataset sintético está disponível neste repositório do \href{https://github.com/dl4nlp-rg/PF08-Gustavo-Paulo}{GitHub} na pasta $date\_text\_norm$ e mais detalhes sobre sua utilização e funcionalidades estão no README do repositório supramencionado.

\subsection{Dataset para Endereços}
O dataset de endereços foi criado com o uso de uma lista de CEPs disponibilizada em \cite{cep}. Com os valores de CEP reais, utilizamos a API disponibilizada por \cite{webmaniabr} para construir um dataset com nomes reais de logradouros, bairros, cidades e estados brasileiros. Estes foram combinados com valores aleatórios de números e formatos de complementos de modo a apresentar $22$ formatos de endereços em português brasileiro. A implementação pode ser encontrada neste repositório do \href{https://github.com/dl4nlp-rg/PF08-Gustavo-Paulo}{GitHub} na pasta $address\_text\_norm$ e mais detalhes sobre sua utilização e funcionalidades estão no README do repositório supramencionado.

\section{Experimentos}
\label{sec:3}
A motivação para a escolha do modelo \texttt{T5} se baseou em três pontos: 
\begin{enumerate}
    \item o modelo é estado da arte em PLN;
    \item o \texttt{T5} possui codificador e decodificador;
    \item é possível utilizar o Google colab no treinamento com a versão do \texttt{T5}-\textit{small}.
\end{enumerate}
O \texttt{T5} é um dos poucos modelos de PLN que a entrada e saída são textos, ao contrário do BERT \cite{bert} que só possui codificador, o \texttt{T5} já inclui a geração de texto na etapa de decodificação. Utilizamos o modelo e tokenizador da \textit{Hugging Face} \cite{hugging_face}, que é a biblioteca  mais famosa para modelos  de PLN estado da arte com arquitetura \textit{transformers}.

Utilizamos a acurácia como métrica de avaliação, onde a predição do modelo precisa ser exatamente igual ao rótulo do dataset, logo qualquer caracter diferente predito é considerado como erro na amostra.
 
\subsubsection{Configuração dos treinamentos}Todos os treinamentos foram realizados no Google colab utilizando GPU e não foi realizado grid-search nos hiperparâmetros. Utilizamos mini-batch de $16$ para os experimentos de datas completas, relativas e endereços e mini-batch de $4$ para datas incompletas. A  \textit{learning rate} $= 5$e$-5$ e otimizador \textit{AdamW} foram usados em todos experimentos. Para os experimentos de datas utilizamos a sequência de entrada e saída de $48$ e $16$ respectivamente. Nos experimentos dos endereços a sequência para entrada e saída foi a mesma  $128$. O código com os notebooks estão no repositório do \href{https://github.com/dl4nlp-rg/PF08-Gustavo-Paulo}{GitHub}, logo a reprodução desses resultados pode ser executada de forma imediata.  

\subsection{Experimentos com Datas Completas}
O dataset utilizado nesse experimento possui a seguinte configuração: $73000$  amostras no intervalo de ano $(1921$ até $2120)$, $34$ formatos utilizados para o treinamento e $11$ formatos utilizados para o teste, escolhidos de forma aleatória. O nível de rúido do dataset foi variado entre o intervalo $(0\%$ até $50\%)$. O modelo também foi avaliado em $50$ datas abaixo e acima do conjunto de treinamento, isto é, datas inferior ao ano $1921$ e acima de $2120$. Fizemos os testes com o dataset em  português Brasil e em inglês, os detalhes dos resultados estão na tabela \ref{complete_date}.

\begin{table}[H]
\centering\centering\resizebox{0.5\textwidth}{!}{
\begin{tabular}{ccccc}
\midrule
\midrule
\textbf{Tipo de}    & \textbf{Nivel de}    & \textbf{Datas no}    & \textbf{Datas Inf.}       & \textbf{Datas Sup.} \\
\textbf{data}       & \textbf{ruído} & \textbf{intervalo} & \textbf{intervalo} & \textbf{intervalo} \\
\midrule
 
                            & --    & 0.98          & \textbf{0.80}  &  0.62 \\
\textbf{\texttt{Datas-PT}}  & 10\%  & 0.97          & 0.71           &  0.60 \\
\textbf{\texttt{Completas}} & 30\%  & \textbf{0.99} & 0.72           &  0.58 \\
                            & 50\%  & 0.98          & 0.77           &  \textbf{0.71} \\
\midrule

                            & --    & \textbf{0.99} & \textbf{0.86}  &  0.74 \\
\textbf{\texttt{Datas-EN}}  & 10\%  & 0.99          & 0.80           &  0.69 \\
\textbf{\texttt{Completas}} & 30\%  & 0.99          & 0.76           &  0.55 \\
                            & 50\%  & 0.99          & 0.82           &  \textbf{0.78} \\
\midrule
\midrule
\end{tabular}
}
\caption{Resultados dos experimentos de datas completas.}
\label{complete_date}
\end{table}

\subsection{Experimentos com Datas Incompletas}
O dataset desse experimento possui a seguinte configuração: para datas do tipo DD/MM existem  $2500$ amostras e para datas MM/AAAA existem $7200$ amostras, ambas no intervalo de ano $(1921$ até $2120)$. De forma similar, foi usado $34$ formatos para o treinamento e $11$ formatos para o teste. O nível de rúido do dataset foi variado entre o intervalo $(0\%$ até $50\%)$. O modelo também foi avaliado em $50$ datas abaixo e acima do conjunto de treinamento, isto é, datas inferior ao ano $1921$ e acima de $2120$. Fizemos os testes com o dataset em português Brasil e em inglês, os resultados estão na tabela \ref{incomplete_date}. 

\begin{table}[H]
\centering\centering\resizebox{0.5\textwidth}{!}{
\begin{tabular}{ccccc}
\midrule
\midrule
\textbf{Tipo de}    & \textbf{Nivel de}    & \textbf{Datas no}    & \textbf{Datas Inf.}       & \textbf{Datas Sup.} \\
\textbf{data}       & \textbf{ruído} & \textbf{intervalo} & \textbf{intervalo} & \textbf{intervalo} \\
\midrule
 
                            & --      & 0.97           & 0.68           &  \textbf{0.53} \\
\textbf{\texttt{Datas-PT}}  & 10\%    & 0.98           & \textbf{0.72}  &  0.40 \\
\textbf{\texttt{Incompletas}} & 30\%  & 0.96           & 0.63           &  0.44 \\
                            & 50\%    & \textbf{0.99}  & 0.65           &  0.42 \\
\midrule
                            & --      & 0.96           & 0.83           &  0.49 \\
\textbf{\texttt{Datas-EN}}  & 10\%    & \textbf{0.98}  & \textbf{0.80}  &  0.45 \\
\textbf{\texttt{Incompletas}} & 30\%  & 0.97           & 0.74           &  0.46 \\
                            & 50\%    & 0.96           & 0.74           &  \textbf{0.51} \\

\midrule
\midrule
\end{tabular}
}
\caption{Resultados dos experimentos de datas incompletas.}
\label{incomplete_date}
\end{table}

\subsection{Experimentos com Datas Relativas}
O dataset utilizado nesse experimento possui $18$ formatos. Foram usados $13$ tipos para o treinamento e $5$ para o teste. O nível de ruído do dataset foi variado em dois cenários: sem ruído e com ruído de $30\%$. Também foram utilizados três tamanhos de datasets com português brasileiro e inglês. Os detalhes dos resultados estão na tabela \ref{relative}. 

\begin{table}[H]
\centering\centering\resizebox{0.5\textwidth}{!}{
\begin{tabular}{ccccc}
\midrule
\midrule
\textbf{Tipo de} & \textbf{Nivel}    & \textbf{Dataset} & \textbf{Dataset} & \textbf{Dataset}  \\
\textbf{Data} & \textbf{de Ruído} & \textbf{1800a} & \textbf{4500a}   & \textbf{9000a}\\
\midrule
\textbf{\texttt{Datas-EN }}  & --    & \textbf{0.98} & \textbf{0.96} & \textbf{0.94} \\
\textbf{\texttt{Relativas}}  & 30\%  & 0.79          & 0.80 & 0.89 \\
\midrule
\textbf{\texttt{Datas-PT}}  & --     & 0.55          & 0.80 & 0.96 \\
\textbf{\texttt{Relativas}}  & 30\%  & 0.33          & 0.69 & 0.80 \\
\midrule
\midrule
\end{tabular}
}
\caption{Resultados dos experimentos de datas relativas. Dataset $1800$a $= 1800$ amostras, Dataset $4500$a $= 4500$ amostras e Dataset $4500$a $= 9000$ amostras.}
\label{relative}
\end{table}

\subsection{Experimentos com Endereços}
Para o experimento de endereços, consideramos entradas de 22 formatos, onde cada formato possui aproximadamente $750$ amostras. A saída esperada possui o seguinte formato: \textit{Rua, número e complemento, bairro, cidade e estado}. Diferente dos experimentos de datas, aqui o dataset só possui está em português brasileiro e o resultado é diretamente afetado pelo nível de ruído, como pode ser visto na tabela \ref{address}.

\begin{table}[H]
\centering\centering\resizebox{0.35\textwidth}{!}{
\begin{tabular}{ccc}
\midrule
\midrule
\textbf{Tipo} & \textbf{Nivel}  & \textbf{Pontuação} \\
              & \textbf{de Ruído} & \textbf{Acurácia}   \\
\midrule
                             & --    & \textbf{0.69}  \\
\textbf{\texttt{Endereços}}  & 30\%  & 0.65  \\
                             & 50\%  & 0.53  \\
\midrule
\midrule
\end{tabular}
}
\caption{Resultados da normalização de endereços.}
\label{address}
\end{table}

\subsection{Experimento Unificado}
Fizemos um experimento unificado com um dataset que contém datas relativas, incompletas, completas e endereços. Esse dataset possui $33039$ amostras com $193$ formatos, o dado não possui ruído e $50\%$ das amostras são de endereços. Utilizamos $48$ formatos distintos escolhidos aleatóriamente para a etapa de validação do experimento e todos os dados estão em português brasileiro. Para este experimento avaliamos a capacidade do modelo de lidar com datas e endereços em uma mesma instância, verificando a eficácia da aplicação do prefixo. Os resultados estão na tabela \ref{unificado}.

\begin{table}[H]
\centering\centering\resizebox{0.45\textwidth}{!}{
\begin{tabular}{ccc}
\midrule
\midrule
\textbf{Tipo} & \textbf{Nivel}  & \textbf{Pontuação} \\
              & \textbf{de Ruído} & \textbf{Acurácia} \\
\midrule
                                     & --    & \textbf{0.79}  \\
\textbf{\texttt{Datas e Endereços}}  & 10\%  & 0.80  \\
\textbf{\texttt{sem prexixo}}        & 30\%  & 0.75  \\
\midrule
                                     & --    & \textbf{0.78}  \\
\textbf{\texttt{Datas e Endereços}}  & 10\%  & \textbf{0.83}  \\
\textbf{\texttt{com prexixo}}        & 30\%  & 0.79 \\

\midrule
\midrule
\end{tabular}
}
\caption{Resultados do experimento unificado.}
\label{unificado}
\end{table}

\subsection{Discussão dos resultados}
\subsubsection{Datas}
Com relação a diferença do resultado entre as línguas português brasileiro ou inglês devemos notar que o \texttt{T5} foi pré-treinado em um corpus de texto em inglês, e os resultados foram melhores na língua inglesa em todos os casos de datas: completas, incompletas e relativas. Com relação ao nível de ruído que simula erros de OCR, o experimento mais sensível ao nível de ruído é o de data relativa, seguido de data completa. Nossa intuição é que o nível ruído está relacionado ao tamanho da sequência, as datas relativas em média possuem maior sequência do que as datas completas, que por sua vez são mais extensas que datas incompletas.

\subsubsection{Endereços}
Acreditamos que o ponto principal para resultado não ter atingido melhor acurácia vem do fato que o \texttt{T5} foi pré-treinado em corpus inglês. Semelhante ao resultado de data relativa que também é diretamente afetado pelo nível de ruído. A solução de normalização de endereço possui uma sequência mais longa do que datas, e uma longa sequência com avaliação da acurácia dificulta bons resultados. Uma possível solução para contornar esse desafio seria o uso da métrica BLEU \cite{BLEU}, mas ainda sim para o contexto de normalização de texto espera-se na saída que os resultados sejam exatamente iguais aos esperados. Por isso a acurácia com correspondência exata continua considerada por nós commo sendo a melhor métrica. 

\section{Conclusão}
\label{sec:4}
Este artigo apresentou uma solução para normalização de datas e endereços que soluciona dois problemas apresentados em métodos com estrutura rígida, tais como o ReGex e DateParser: generalização e tolerância a dado com ruído.  Utilizamos um modelo de PLN com rede neural profunda$-$\texttt{T5} que contém na versão \textit{small} $65$ milhões de parâmetros. Esse grande modelo traz um bom poder de generalização e tolerância a um dado com ruído.

Além disso, também foi desenvolvido dois datasets sintéticos, um para cada problema que são flexíveis na escolha do tamanho, da língua e no intervalo de interesse do dado. Com relação aos experimentos: em datas utilizamos as línguas português brasileiro e inglesa.  A motiviação para o  desenvolvimento do dataset em inglês é confirmada pelos melhores resultados dos experimentos de normalização de datas. Para o experimento de endereços notamos uma menor acurácia, entretanto esta tarefa de normalização requer uma saída canônica, o que torna mais difícil a avaliação por acurácia levando em conta a correspondência exata. Por último, o experimento unificado é relevante no sentido que comprova que o modelo desenvolvido neste artigo é robusto, pois com mais de $190$ tipos de formatos, nossa proposta conseguiu acurácia acima de $75\%$ considerando endereços e datas.

A reprodutividade sempre foi ponto chave dessa pesquisa, apesar de utilizarmos GPU para os experimentos, utilizamos GPU gratuitas do Google Colab e todo código com os datasets estão disponíveis em um repositório github. Também utilizamos o PyTorch Lightning para reprodução mais facilmente do código e gostaríamos de agradecer ao \texttt{Israel Campiotti} por dicas valiosas e um template refinado de PyTorch Lightning.

\section{Trabalhos Futuros}
\label{sec:5}
Pretendemos refazer alguns experimentos com um \texttt{T5} treinado em português, acreditamos que existirá uma melhor relação de comparação entre a língua inglesa e portuguesa. Também desejamos incluir mais dicionários de diferentes línguas em nossos datasets, possibilitando que outros grupos de pesquisa utilizem essa solução.

%\end{multicols}{2}

\end{document}